\RequirePackage{fix-cm}

\documentclass[envcountsect,twocolumn]{svjour3}

\smartqed

\usepackage{nicefrac}
\usepackage{lipsum}
\usepackage{amssymb} 
\usepackage{adjustbox}
\usepackage{scalefnt}
\usepackage{graphicx,url}
\usepackage[utf8x]{inputenc} 
\usepackage{float}
\usepackage{subfigure}
\usepackage{xcolor,colortbl}
\usepackage{dblfloatfix}
\usepackage{array,multirow,graphicx}
\usepackage{amsmath}
\def\:{\hskip0pt}

\journalname{Preprint}

\begin{document}

\title{Modelling Energy Consumption based on Resource Utilization}

\author{Lucas Venezian Povoa \and Cesar Marcondes \and Hermes Senger}

\institute{
   L. Venezian Povoa \at
   Federal Institute for \\
   Education, Science, and Technology of São Paulo\\
   Caraguatatuba, São Paulo -- Brazil\\
   \email{venezian@ifsp.edu.br}
   \and
   L. Venezian Povoa \at
   Aeronautics Institute of Technology\\
   São José dos Campos, São Paulo -- Brazil
   \and
   C. Marcondes \and H. Senger \at 
   Computer Science Department, \\
   Federal University of São Carlos\\
   São Carlos, São Paulo -- Brazil\\
   \email{\{senger, marcondes\}@dc.ufscar.br}
}

\date{\today}

\maketitle

\begin{abstract}
Power management is an expensive and important issue for large computational infrastructures such as datacenters, large clusters, and computational grids. However, measuring energy consumption of scalable systems may be impractical due to both cost and complexity for deploying power metering devices on a large number of machines. In this paper, we propose the use of information about resource utilization (e.g. processor, memory, disk operations, and network traffic)  as proxies for estimating power consumption. We employ machine learning techniques to estimate power consumption using such information which are provided by common operating systems.
Experiments with linear regression, regression tree, and multilayer perceptron on data from different hardware resulted into a model with 99.94\% of accuracy and 6.32 watts of error in the best case.

\end{abstract}

\section{Introduction}\label{sec:introduction}

Over the years, managing energy efficiency of Information and Communication Technologies (ICT) has increasingly emerged as one of the most critical environmental challenges. Due to ever increasing demand for computing resources, emissions footprint, increased energy price and tougher regulations, improving energy efficient became priority for datacenters, especially to the massive ones. This concern is pervasive in ICT, from development of more energy efficient devices to greener virtualization, resource consolidation, and, finally, definition of new architectures, services, and best practices.

In 2007, a Gartner's Report showed that ICT industry generated 2\% of global CO2 \cite{gartner_co2} emissions. From which, 23\% came from datacenters. A Greenpeace's report \cite{greenpeace} stated that ``datacenters are the factories of the 21st century in the Information Age'', however, they can consume as much electricity as 180,000 homes. 

The constant reduction in computation resources prices, accompanied with popularization of on-line businesses, and wide spread of Internet and wireless networks, lead to the rapid growth of massive datacenters, consuming large amounts of energy. Indeed, nowadays, datacenters that execute Internet applications consume around 1.3\% of the energy produced in the world~\cite{gao2012s}. It is expected in 2020 that this amount will rise to near 8\% \cite{koomey2011growth}. 
In such scenario, improving power efficiency on ICT installations and datacenters is mandatory. To overcome this challenge, several strategies have been proposed, such as resource consolidation  \cite{orgerie_et_al_2010,lee_e_zomaya_2012,song_et_al_2009}, and improving resources utilization \cite{barroso_e_holzle_2007}. 
 
In general, better energy efficiency can be achieved by means of actuation strategies which need the continuous power consumption measurement. The deployment of power meters may be prohibitive in terms of cost in datacenters with many thousands of computers. Furthermore, external metering instruments require physical system access or invasive probing \cite{song2013simplified}, which can be not avaliable. On the other hand, software estimators for power consumption can be easily deployed at almost negligible cost. 

An usual approach is to use internal performance counters provided by the hardware \cite{naveh2011power} and by the operating system to derive models that estimate power consumption \cite{contreras2005power,joseph2001run,isci2003runtime,bertran2010decomposable}. Such models can be used by on-the-fly power saving strategies which need continuous power consumption estimation. Other possible applications include simulators that evaluate the power consumption of workloads based on performance and resource usage counters  (e.g., register file usage, number of page faults, number of I/O operations per second).

%%%%%%%%%%%%%%%%%%%%%%%%%%%%%%%%%%%%%%%%%%%%%%%%%%%%%%%%%%%%%%%%%%%%%%%%%%%%%
% Contributions
%%%%%%%%%%%%%%%%%%%%%%%%%%%%%%%%%%%%%%%%%%%%%%%%%%%%%%%%%%%%%%%%%%%%%%%%%%%%%

In this paper, we propose models that use counter of both performance and resource utilization as proxies for power consumption. Differently from most of the related work, our models are not limited to predict power consumption of specific components, but of whole machine.

We assume a good model should include all performance counters which significantly influence the power consumption. However, the excess of parameters and non-linear relations between these variables and power consumption can produce complex and inaccurate models. Having this on mind, we also investigate which operating system counters can be used to build robust and accurate models.

In a previous work \cite{povoa2013model}, we studied the correlation between a set of resource utilization counters provided by an operating system and the power consumption on a typical server machine. We also proposed a first-cut linear regression model with encouraging results.

Now, we further elaborate on correlation analysis and estimation of power consumption from resource utilization variables (i.e., counters) provided by operating systems. With this purpose, we apply nine models based on ($i$) Multiple Linear Regression (MLR), ($ii$) Regression Tree (RET), and ($iii$) Multilayer Perceptron (MLP), an  Artificial  Neural Network (ANN) which are experimentally evaluated on two different hardware \footnote{The models were implemented in R (using RSNNS) and they are available under the GNU General Public License version 3 at \url{https://github.com/lucasvenez/ecm} along with the employed dataset}.

The remainder of this paper is organized as follows. Section~\ref{sec:methods} describes the modeling approach, the workload, and the testbed used for experiments. Section~\ref{sec:var} shows the collected variables and which of them have most impact on power consumption to be considered in the models. Section~\ref{sec:models} describes the power consumption models based on the MLR, RET, and MLP methods. Section~\ref{sec:performance-accuracy-analysis} presents accuracy and performance analysis of each proposed model. Section~\ref{sec:related_work} presents some related work. Finally, Section~\ref{sec:conclusion} presents our final remarks.

\section{Materials and Methods}\label{sec:methods}

This paper aims to provide a characterization of the power consumption for a wide variety of machines. With this purpose, we propose new models which provide accurate estimates for the power consumption based on resource utilization measurements commonly supported by the operating system used from commodity computers to datacenter servers. In this section, we describe the initial steps to develop such models.

%\subsection{Background}\label{sec:definitions}

%Computer components  consume energy to produce work, whereas power is the  rate at which a device consumes electricity, measured  in joules per second or watts (W). In a datacenter, power is typically reported in kilowatts (kW). On the other hand, energy is the total amount of  electricity consumed over some time interval. Thus, energy is power rate integrated over time, expressed in kilowatt-hour (kWh). 

%It is also worth noting that power is typically measured not by taking an instantaneous measurement, but instead, by measuring energy consumption over a period of time and then dividing by that time period to compute an average power. The average is used because instantaneous power at a device can vary over time in a cyclic fashion 50 to 60 times per second. Thus average power becomes more meaningful  because it is stable and reflects an average rate of consumption over a time interval.

\subsection{Modelling Approach}\label{sec:approach}

\begin{figure}[h!]
\includegraphics[width=8.0cm]{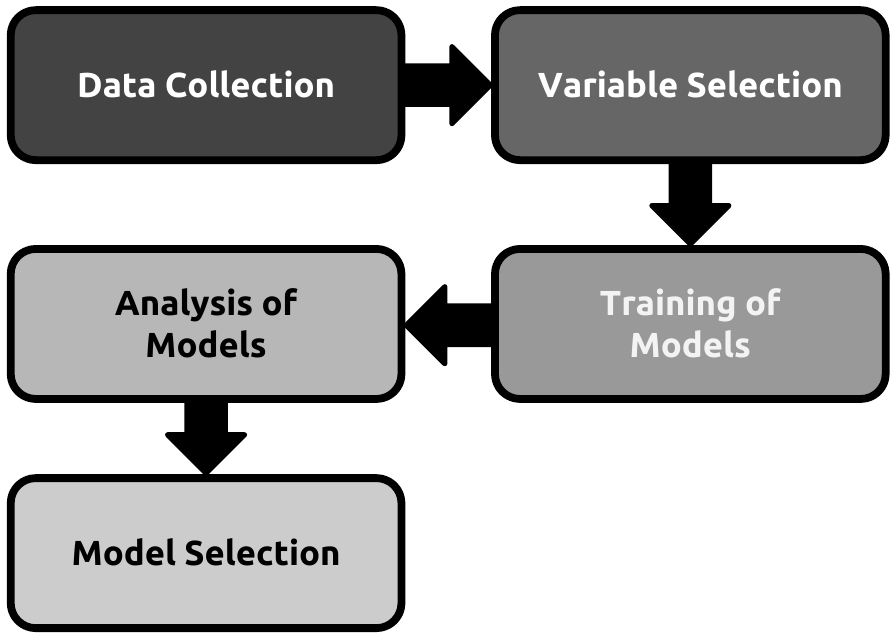}
\caption{Power Consumption Modeling Approach.}
\label{fig:steps}
\end{figure}

In order to model power consumption for different computers, we employed a five-step method as depicted in Figure~\ref{fig:steps}. 

In {\it Data Collection} step, a synthetic workload is executed and an agent is used to collect data about resource utilization from the operating system \cite{iaee,iaee_impl}. The agent captures forty-seven variables from \emph{/proc} directory. Second step is the {\it Variables Selection}, which aims to select variables that are influential to power consumption. We employed a correlation method called Maximal Information Coefficient (MIC)~\cite{mine} that evaluates the correlation of a pair variables regardless of the distribution. Next step, called {\it Training of Models}, aims to fed models with resource utilization samples and reads of the actual energy consumption measured in the testbed. {\it Analysis of Models} is the next step, in which focus on evaluating models accuracy through a set of different metrics. The final step is {\it Model Selection}, where the best estimation model for power consumption is selected.

\subsection{Data Collection}\label{sec:workload}

Our objective is to characterize power consumption using operating system counters as proxies for energy consumption. We built a synthetic workload instead of using real applications or benchmarks aiming to conceive  energy consumption models which are suitable for any application, while avoiding collinearity problems which may compromise regression models \cite{pazzani1999independent,bertran2010decomposable}.

The workload was designed to avoid cross dependency among the features fed to the model and produce all power consumption states for all system components such as memory, hard disk, processor, network interfaces and I/O \cite{da2010methodology}. Workload was produced by using three open source tools: ($i$) \textit{stress} \cite{stress} was used to produce utilization of resources such as processor, memory, hard disk and input and output;  ($ii$) \textit{cpulimit} \cite{cpulimit} was used to generate random periods of idleness to produce several levels of processor utilization; and ($iii$) \textit{iperf} \cite{iperf} to generate network traffic.

Workload was produced with the following characteristics. CPU utilization varied between 0\% and 100\% in several cycles, being increased in steps of 5\% each. Each experiment was composed by $P_i = 2i - 1$ processes with $1 \leq i \leq N_{cpu}$, where $N_{cpu}$ is the number of processors in the machine for the  $i-th$ test. Memory utilization ranged from 512MB to the physical memory size. For the $i-th$ experiment, one application process allocates  $M_i = 256(i + 1)$ MB of memory, such that $1 \leq i \leq \frac{M_{size}}{256} - 1$. Hard disk utilization varied from 1GB to 64GB, being produced by one process. For each experiment, the amount of disk space allocated is $C_i = 2i-1$ GB, where $1 \leq i \leq 17$.

I/O workload was expressed by the number of processes that performed the message exchanges between the memory and the hard disk. The amount of processors exchanging messages was given by $P_i = 10i$, with $1 \leq i \leq 10$, where $i$ is the number of the experiment.

In the beginning, only one parameter was selected to vary for each experiment, in order to capture its influence on power consumption. Then, parameters were varied to test every all-to-all combinations of the several parameter levels, in order to capture their influence on power consumption and as well as parameter interactions.  For each combination of parameters and level, the workload is executed for two minutes. The overall experiment took about fifteen hours to be carried out, producing about 51,000 entries, each entry  containing measures from 29 variables of resources utilization and the power consumption.

\subsection{Testbed Used for Experiments}\label{sec:testbed}

\begin{figure}[ht!]
\centering
\includegraphics[width=8.5cm]{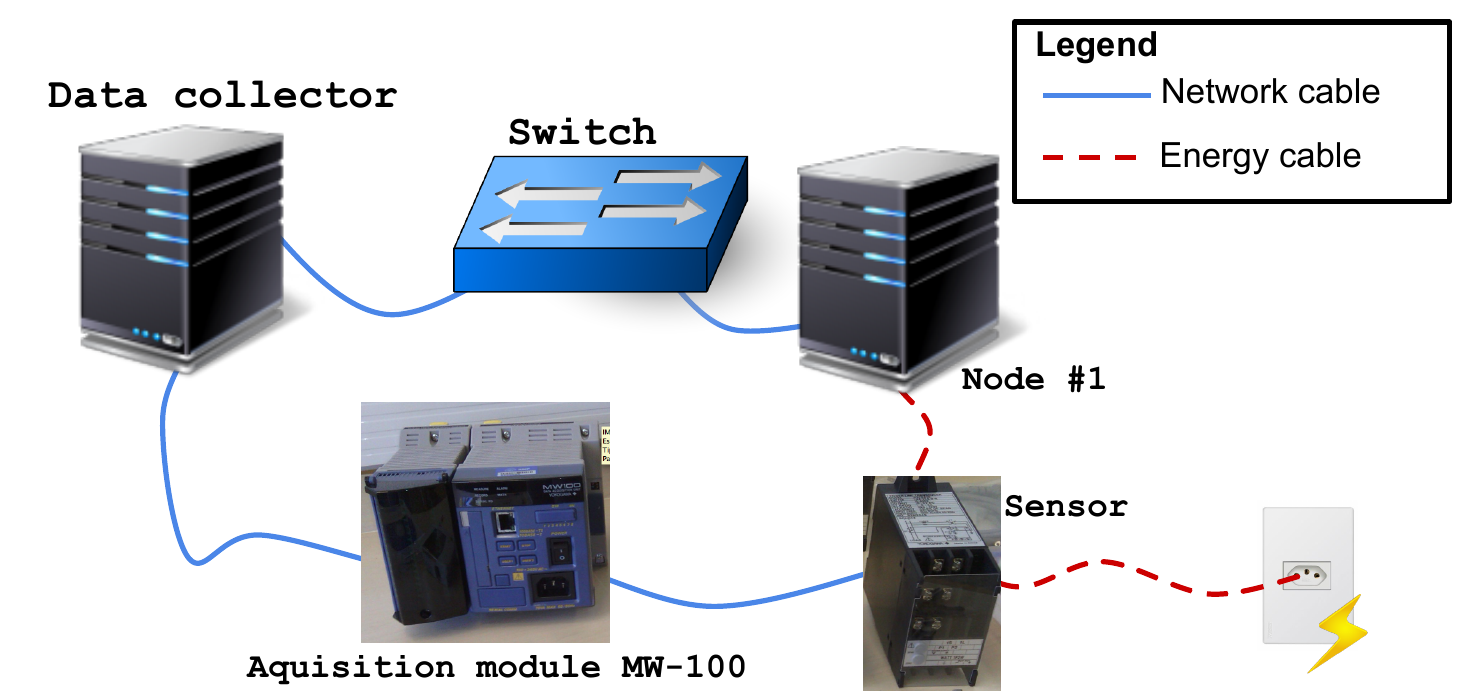} 
\caption{Experiment environment with a node, an energy consumption meter, a module and a data storage.}
\label{fig:environment}
\end{figure}

The testbed used for experiments is depicted in Figure~\ref{fig:environment}. Some cluster nodes were instrumented to measure power consumption while running workloads. 
We employed two nodes with different architectures in the experiments, which have their hardware configuration summarized in Table~\ref{tab:arch}.

In order to obtain precise power consumption measures, we used a power sensor Yokogawa model 2375A10 \cite{2375a10}. This device works connected to the power supply, and provides data to one data acquisition module model MW-100-E-1D \cite{mw100}. The acquisition module probes and saves measures on power consumption in watts every 100 milliseconds. A software agent was implemented for collecting data from the acquisition module via a network interface using the telnet protocol.

\begin{table}[h!]
\centering
\scalefont{1.0}
\caption{Hardware configurations with one 1Gbps network interface running Ubuntu 11.10 kernel 3.0.0-12.}

\begin{tabular}{lrr}

\hline 

\multicolumn{1}{c}{\textbf{Hardware}} & 
\multicolumn{1}{c}{\textbf{A1}}    & 
\multicolumn{1}{c}{\textbf{A2}}      \\

\hline

\textit{Processor model}   & 
Intel Core i5-2400 &
AMD Opteron 246  \\

\textit{Cores} &
4              &
2              \\

\textit{Frequency} &
3.10 GHz           &
2.00 GHz           \\

\textit{Memory} &
4GB SDRAN       &
8GB SDRAN       \\

\textit{Disk}    &
1 $\times$ 500GB &
4 $\times$ 240GB \\

\hline

\end{tabular}
\label{tab:arch}
\end{table}

\section{Variables Selection}\label{sec:var}

The design of accurate models depend upon a good selection of resource utilization counters \footnote{Details about each independent variable considered in this paper are described in Supplementary Material.} which present significant influence on power consumption and do not produce noise.

\begin{figure*}[b!]
\includegraphics[width=\textwidth]{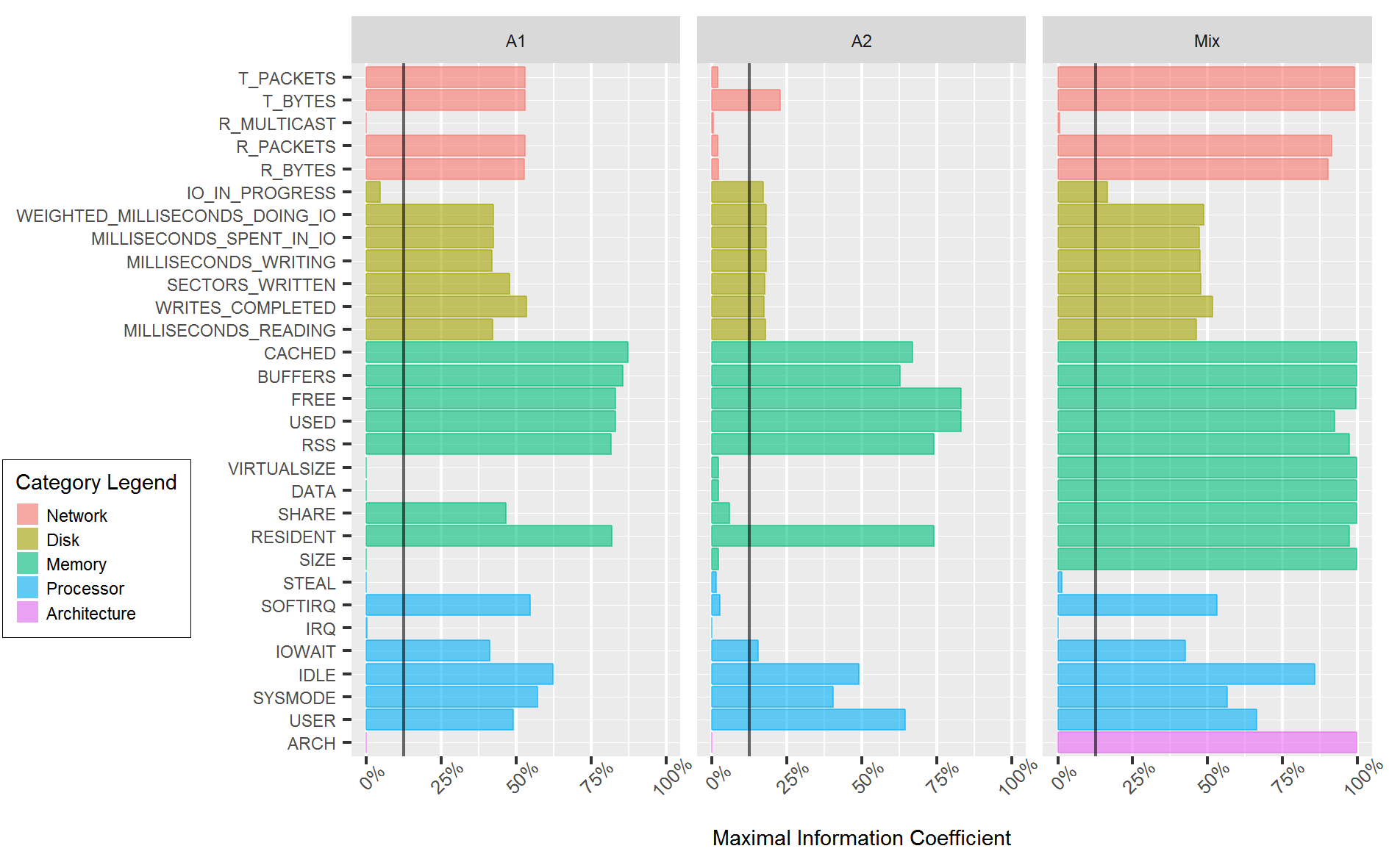}
\caption{Maximal Information Coefficient for the datasets of each architecture and for the mixture thereof.}
\label{fig:micresults}
\end{figure*}

For the sake of clearness and understandability, a model for estimating energy consumption should be simple, i.e., to consider only a subset composed of the most influential variables on energy consumption. With this purpose, we identified from the set of observed variables, the subset with the highest correlation with the dependent variable (i.e., the energy consumption).

In order to evaluate correlation among variables, two main criteria should be considered. The first on is generality, which  means capacity of identifying any relation type, not limited to specific types of correlation functions such as linear, exponential and periodic correlations, for instance. The later is equitability, that means the ability to provide a unique index to express relation with the same noise level, even for functions of different types. 

Consequently, we chose  a statistical method MIC to discover what are the most impacting variables for energy consumption. It is part of a set of tools named Maximal Information\:-\:based Nonparametric Exploration (MINE)~\cite{mine}. As result, MIC produces values between 0 and 1, where zero means absence of correlation between the pair of variables and 1 means full correlation.

Each architecture generated a dataset of the variables described previously. Based on these two datasets was generated a third one containing all data of both architectures, called Mix, joined to another variable, which determines whether the data is related to the architecture A1, described as -1, or to the architecture A2, described as 1.

Figure~\ref{fig:micresults} shows results of the MIC between each independent variable, i.e., operating system's variables, and the dependent variable, i.e., energy consumption. In the chart the vertical black line represents the threshold of 10\%, which was applied with the purpose of finding a reduced set of the most impacting coefficients and produce a model with good understandability.

\subsection{Dependent variable analysis}\label{sec:depvaranalysis}

The distribution followed by the dependent variable of a model, which is the energy consumption our case, defines the method that can be used for modeling its behavior. The conventional Linear Regression method, for example, is not appropriated for modeling dependent variables that has no a Gaussian distribution.

\begin{figure}[!htb]
\includegraphics[scale=1]{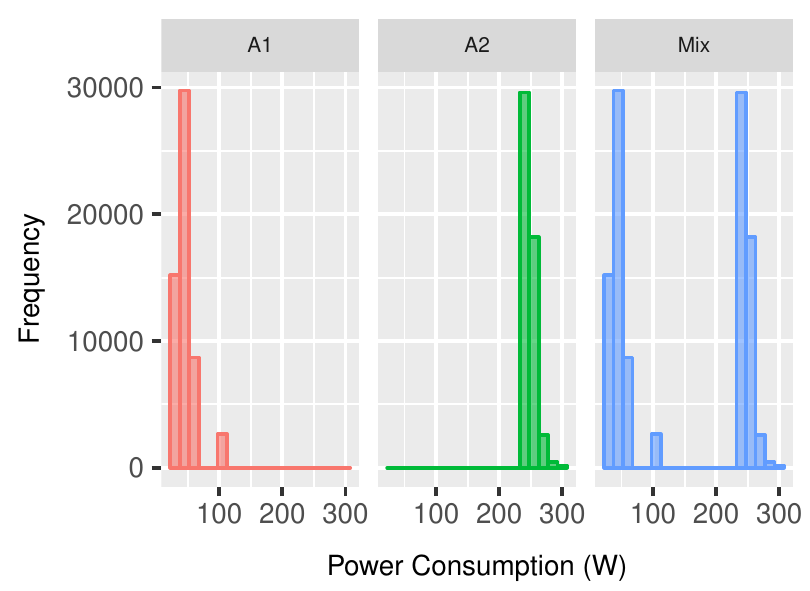}
\caption{Energy consumption histogram.}
\label{fig:power_histogram}
\end{figure}

Figure~\ref{fig:power_histogram} shows the histogram of the energy consumption of each architecture, A1, A2, and Mix. Visually it is noteworthy that neither architecture have a normal distribution.

In order to confirm that power consumption do not present a Gaussian distribution we applied the  Kolmogorov-Smirnov test \cite{ks-test} for each architecture (A1, A2, and Mix). These tests resulted in \textit{p-values} less than $2.2e-16$, which confirm that the dependent variable has no a Gaussian distribution considering a significance level of 5\%.

The datasets A1, A2, and Mix present an average energy consumption (watts) of $46.08$, $249.23$, and $142.62$, respectively.  Their standard deviation are $15.26$, $7.74$, and $102.19$. It is noteworthy that the architectures A1 and A2 have an stable energy consumption but in different ranges as also showed in Figure~\ref{fig:power_histogram}.

\section{Models for Power Consumption}\label{sec:models}

In this section we describe three models for estimating power consumption based on the most influential variables described in the previous Section.

\subsection{Multiple Linear Regression Model}\label{sec:mlr}

A MLR is a type of regression analysis that maps a set of input values $\boldsymbol{X}$ to a response value $y$, requiring that $\boldsymbol{y} \sim N(\mu, \sigma^2)$. However, as show in Section~\ref{sec:depvaranalysis}, neither datasets follow a normal distribution.

For dealing with this divergence, we consider the Central Limit Theorem (CLT), which states that when the size of a given sample increases, the sampling distribution of its average or sum tends to a normal distribution \cite{clt}. Hence, the CLT justifies modelling the energy consumption with the MLR defined as

\begin{equation}
\hat{\gamma} = \alpha + \boldsymbol\beta\boldsymbol{x} + \epsilon
\label{eq:model}
\end{equation} where 
\begin{itemize}
\item $\hat{\gamma}$ is the estimated value of the energy consumption;
\item $\alpha$ is the intersection point of the line of adjustment with the ordinate; 
\item $\boldsymbol\beta$ is the regression coefficients vector; 
\item $\boldsymbol x$ is the independents variables vector; and 
\item $\epsilon$ is the average random error.
\end{itemize}

This model employs the least-squares method for estimating the vector of coefficients $\boldsymbol{\beta}$, which is defined as

\begin{equation}
\boldsymbol{\beta} = (\boldsymbol{X}^{T}\boldsymbol{X})^{-1}\boldsymbol{X}^{T}\boldsymbol{y}
\label{eq:ls}
\end{equation} where

\begin{itemize}
\item $\boldsymbol{X}$ is the matrix of independent variable values; and
\item $\boldsymbol{y}$ is the array of dependent variable values.
\end{itemize}

\begin{figure}[h!]
\centering
\subfigure{\includegraphics[scale=0.8]{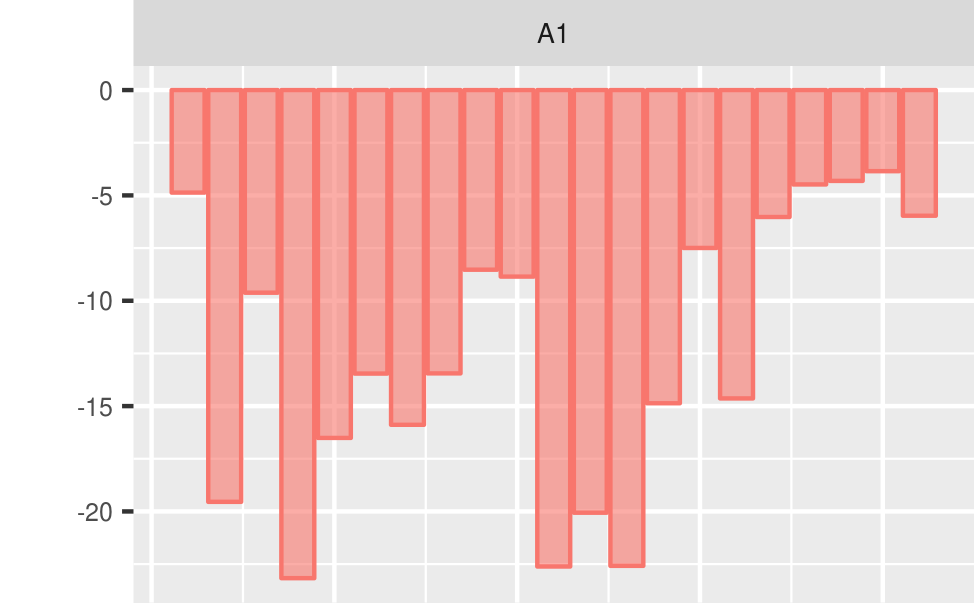}}
\subfigure{\includegraphics[scale=0.8]{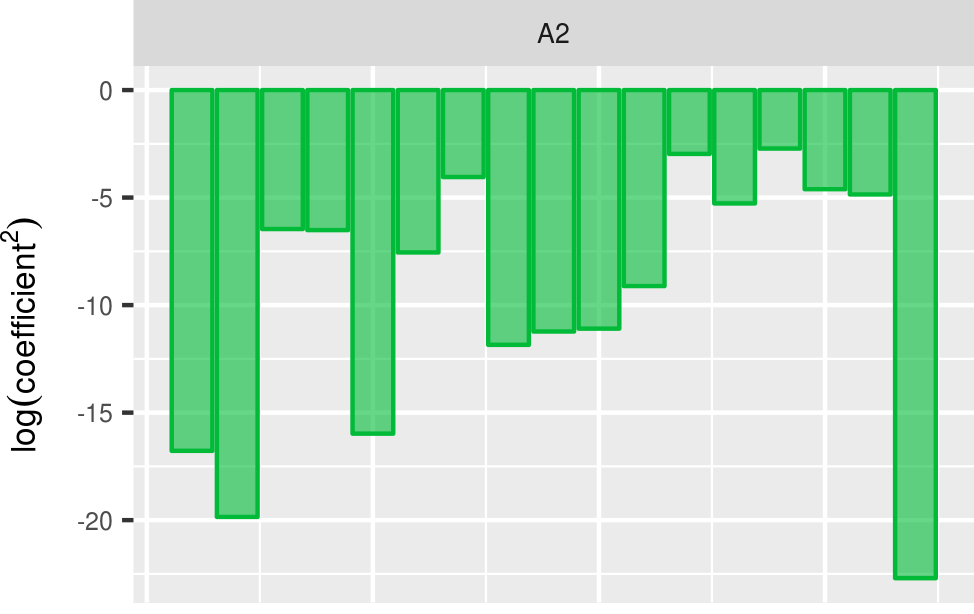}}
\subfigure{\includegraphics[scale=0.8]{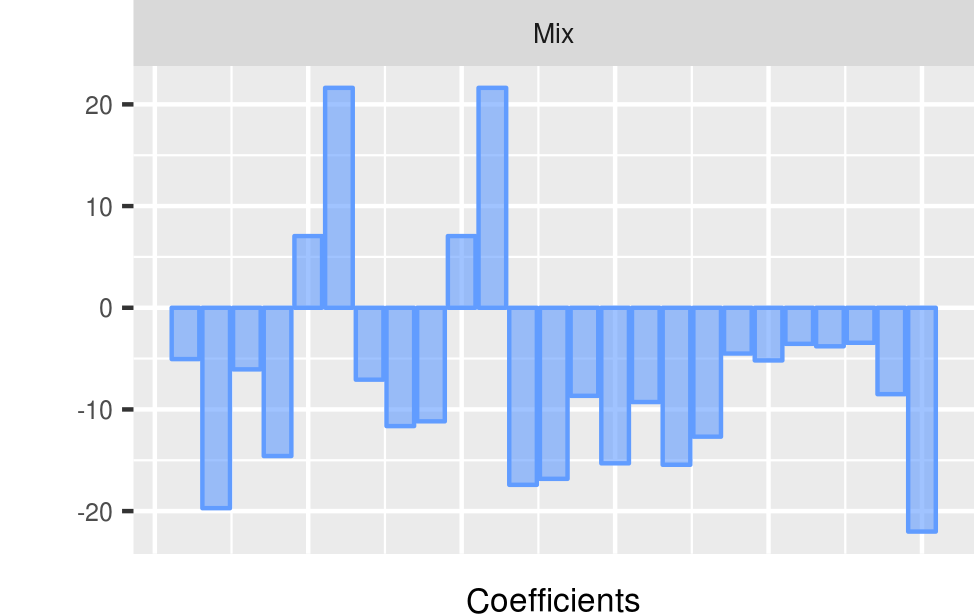}}
\caption{Coefficient values of the MLR models.}
\label{fig:mlrcoeff}
\end{figure}

Figure~\ref{fig:mlrcoeff} shows the model's coefficient values for each architecture under analysis after applying Equation~\ref{eq:ls}. The models for A1, A2, and Mix architectures have the intercept values $-35.35$, $-4001.74$, and $7.9 \times 10^8$, respectively.

Despite the correlation analysis demonstrate high relation between the dependent variable and the independent variable labeled ARCH, the MRL method cannot incorporate this former into the resultant model for the Both dataset.

Despite of the high correlation analysis between the dependent variable and the independent variable ARCH, the MRL method cannot incorporate this former into the resultant model for the Both dataset.
This is limitation for the MLR method for generating a global energy consumption model. A detailed analysis is presented in Section~\ref{sec:accuracy}.

%------------------------------------------------------------------------
\subsection{Regression Tree Model}\label{sec:rt}

A Decision Tree (DT) has a structure composed by leaves, branches and nodes aiming to define a nonlinear predictive model. A RET is a particular case of a DT, where values of dependent variables are continuous. Using a RET as predictor requires a sample be dropped down via the tree until a leaf, which returns the average of its values of the dependent variable~\cite{CART}.

A RET defines its configuration by splitting a node $p$ into two children nodes $l$ and $r$. The split criterion $I(p)$ of $p$ that is used to defined which variables gives the best split, is based on the sum of squared errors $e = \sum_i y_i - \bar y$ and defined in Equation~\ref{eq:split-criterion}~\cite{intro_rpart}.

\begin{equation}
I(p) = e_p - (e_l + e_r)
\label{eq:split-criterion}
\end{equation}

The tree stops to grow when the complexity index $\beta_p$ of a node $p$ is less or equals to a threshold $\alpha$. The complexity index $\beta$ of a node $p$ is calculated as follows:
\begin{equation}
%\textfractionsolidus
%\beta_p = \frac{I(p)}{(n_l + n_r + 1)}
\beta_p = \nicefrac{I(p)}{(n_l + n_r + 1)}
\label{eq:complexity}
\end{equation}
where $n_l$  and $n_r$ are number of elements in the children nodes $l$ and $r$, respectively.

\begin{figure}[h!]
\centering
\subfigure[fig:rtconfa][A1]{\includegraphics[scale=.9]{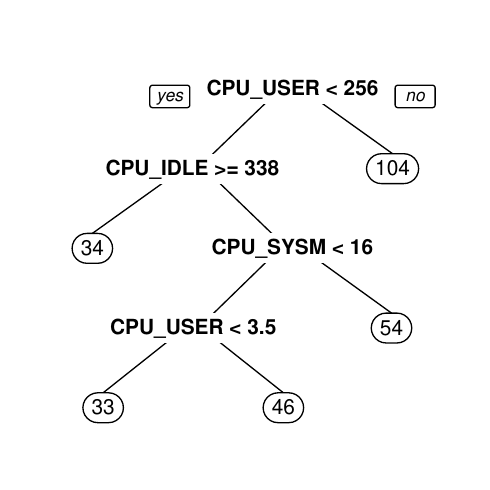}}
\subfigure[fig:rtconfb][A2]{\includegraphics[scale=.9]{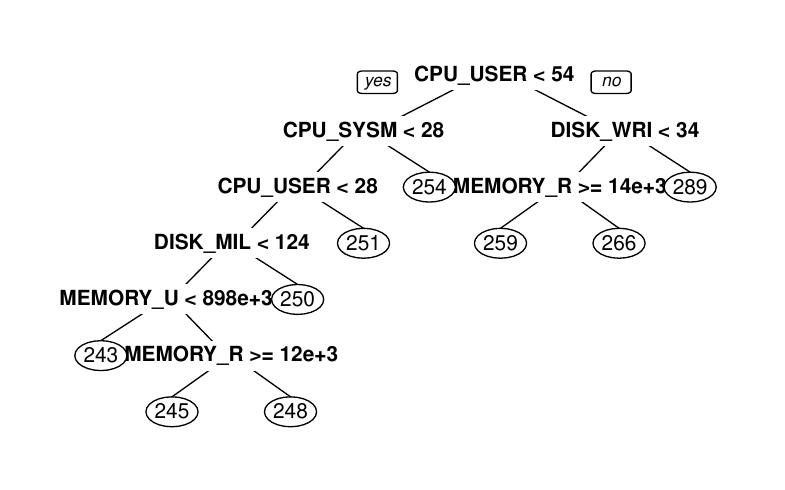}}
\subfigure[fic:rtconfc][Mix]{\includegraphics[scale=.9]{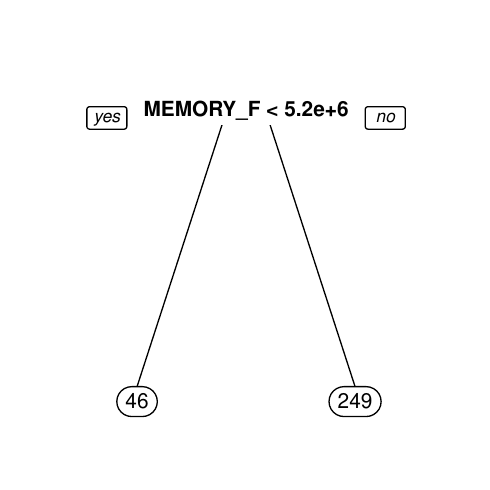}}
\caption{Regression Trees to Forecast Energy Consumption.}
\label{fig:rtconf}
\end{figure}

For the experiments the threshold $\alpha$ was set as $1\%$. Applying our dataset of energy consumption in RETs generated the model illustrated in Figure~\ref{fig:rtconf}.

RET models are easy interpreted, but our results show that important variables are excluded for the model, which evidence a limitation of this method for modeling energy consumption. The resultant model for the Mix dataset represents our worst model, which considers only one independent variable for defining itself.

\subsection{Multilayer Perceptron Model}\label{sec:mlp}

A MLP is an Artificial Neural Network model that maps a set of input values into a set of output values~\cite{mlp} after a  learning process. Recently, it is being successfully applied in different areas, e.g.,  Biometrics \cite{mlp_biometric}, Thermal Engineering \cite{mlp_thermal}, Ocean Engineering \cite{mlp_ocean}, Climatology \cite{mlp_climatology}.

\begin{figure}[h!]
\centering
\includegraphics{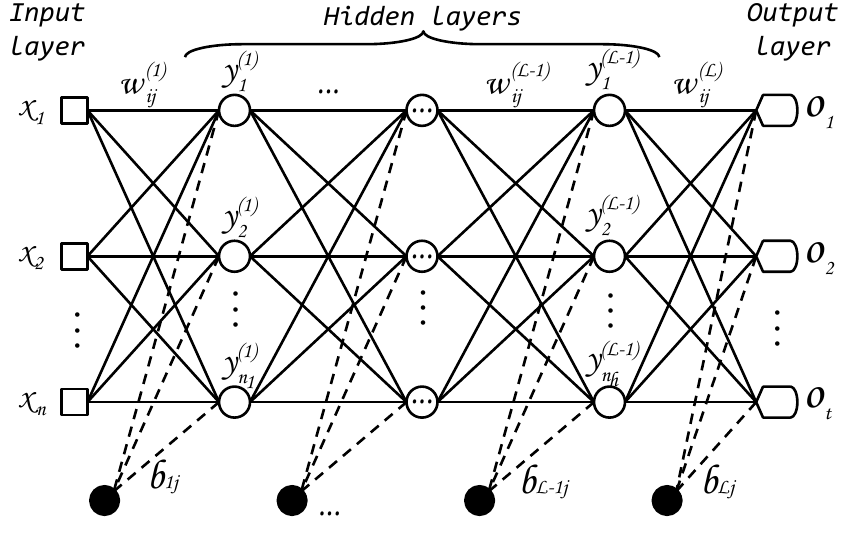}
\caption{Multilayer Perceptron architecture.}
\label{fig:mlp}
\end{figure}

The MLP is composed by an input layer with $n$ sensory units, $h$ hidden layers with $n_h$ neurons each, and an output layer with $t$ neurons. A MLP has $L$ layers, excluding the input layer, and its input values are propagated layer-by-layer. Figure~\ref{fig:mlp} shows the general structure of this model.

The neuron $j$ at layer $l$ has an induced local field $v_j^l$ defined by:
\begin{equation}
v_j^l = \sum_{i=1}^{q^{l-1}}\bigg[w_{ij}^{l} \times y_i^{l-1}\bigg] + b_{lj},
\label{eq:induced-local-field}
\end{equation} where 
\begin{itemize}
\item $q^{l-1}$ is the number of neurons at layer $l-1$;
\item $w_{ij}^{l}$ is the weight between the neuron $i$ at layer $l - 1$ and the neuron $j$ at layer $l$;
\item $y_{i}^{l-1}$ is the value of the neuron $i$ at antecedent layer $l-1$; and
\item $b_{lj}$ is a bias at layer $l$ connected to neuron $j$.
\end{itemize}

Each neuron is a computational unit composed by an input signal, a weight, and a nonlinear activation function. The output value $y_j^l$ for neuron $j$ at layer $l$ is denoted by

\begin{equation}
y_j^{l} = \varphi(v^l_j)
\label{eq:mlp_body}
\end{equation} where $\varphi$ is an activation function.

If neuron $j$ is located at layer $l = 0$, its output value $y^0_j = x_j$. The output value $y^L_j$ for neuron $j$ at layer $l = L$ is denoted thought the variable $o_j$.

The MLP's learning process can be supervised or unsupervised. Once we collected both the input and output variables, this research applied the supervised learning process, i.e. a process that uses the expect output for correcting the model's weights. The supervised learning process was performed with the backpropagation algorithm with chunk update, which has the following steps:

\begin{enumerate}
\item forward step, where a set of input values is provided to the sensory units, and its effect is propagated layer-by-layer; and
\item backward step, where the weights are adjusted in accordance with an error-correction rule respecting the Mini-Batch Stochastic Gradient Descent method \cite{minibatch} after $p$ (chunk size) executions of the forward step.
\end{enumerate}

The forward step applies Equation~\ref{eq:mlp_body} for each neuron $j$ in each hidden layer $l$. After that, an error signal $e_j$ related to neuron $j$ at output layer $L$ is denoted by:
\begin{equation}
e_j = d_j - o_j,
\label{eq:mlp-error}
\end{equation} 
where $d_j$ is the $j^{\text{th}}$ element of the desired response vector.

The backward step starts by computing the local gradient $\delta^{L}_j$ related to neuron $j$ at output layer $L$ according to:

\begin{equation}
\delta^{L}_j = e^{L}_j \times \varphi'(v^{L})
\label{eq:gradient}
\end{equation}

When neuron $j$ is located at a hidden layer $0 < l < L$, the local gradient $\delta^{l}_j$ related to neuron $j$ at hidden layer $l$ is defined in by

\begin{equation}
\delta^{l}_j = \varphi'(v^{l}_j) \times \sum_{k=1}^{g}[\delta^{l+1}_{k} \times w_{kj}^{l+1}],
\label{eq:mlp-hidden-local-gradient}
\end{equation} where
\begin{itemize}
\item $\varphi'$ is the derivative of $\sigma$; and
\item $g$ is the number of neurons at layer $l+1$.
\end{itemize}

The new value for the weight $w_{ij}^{l}$ at layer $l$ is defined according to
\begin{equation}
w_{ij}^{l}(n+1) = w^{l}_{ij}(n) - \nicefrac{\eta}{p} \times \sum_{m}^{n}[\delta^{l}_{j}(m) \times y_i(m)],
\label{eq:mlp-weight-update}
\end{equation} where 
\begin{itemize}
\item $n$ is the iteration number, such that $n\mod p = 0$ or $n = N$;
\item $m = \begin{cases}
   n - p + 1, \text{ if } n\mod p = 0\\
   n - (n\mod p - 1) \times p, \text{ if } n = N
\end{cases}$;
\item $p$ is the chunk size; and
\item $\eta$ is the learning-rate.
\end{itemize}

For setting the MLP's configuration for each architecture, we applied an empirical method that consists of:

\begin{enumerate}
\item selecting a random and non-sequential subset of registers from our sample, approximately $20\%$ of all registers;
\item starting the model weights with a random Gaussian distribution with values between $0$ and $1$;
\item ranging the number of hidden layers from $1$ until a descendant precision of the model;
\item ranging the number of neurons at each layer from $\frac{v}{10}$ to $2 \times v$, where $v$ is the number of independent variables at the model;
\item ranging the learning-rate from $0$ to $10$ by $0.25$; and
\item calculating the model precising using the Coefficient of Determination R$^2$ metric for each possible configuration considering previous steps.
\end{enumerate}

After finding the better combination for each architecture, where better combination refers to the one that results in the greatest R$^2$, we applied the full sample for the learning process. In our study, the better configuration for the MLP consists of 3 hidden layers, where each one has the number of nodes equals to double of the number of input variables, a learning rate $\eta = 5$, and a chunk size $p = 50$. As activation function $\varphi$ we used the $tahn$ function defined in Equation~\ref{eq:activation-function}.

\begin{equation}
\varphi(x) = tanh(x) = \nicefrac{(e^{2x} - 1)}{(e^{2x} + 1)}
\label{eq:activation-function}
\end{equation}
where $e \simeq 2.71828$ is the Euler's number.

\begin{figure*}[h!]
\includegraphics[width=\textwidth]{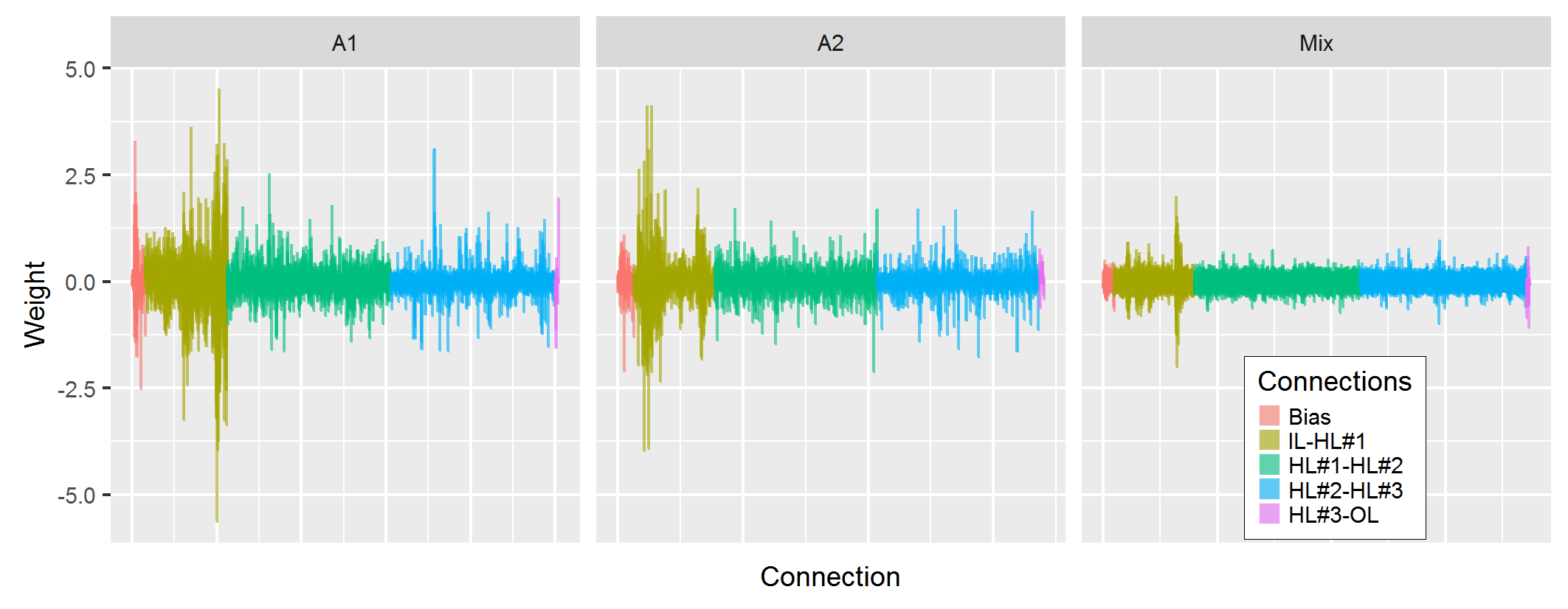}
\caption{MLP's weights for each architecture split into connection categories: Bias, input layer to hidden layer \#1 (IL-HL\#1), hidden layer \#1 to hidden layer \#2 (HL\#1-HL\#2), hidden layer \#2 to hidden layer \#3 (HL\#2-HL\#3), and hidden layer \#3 to output layer (HL\#3-OL).}
\label{fig:mlpweigths}
\end{figure*}

Figure~\ref{fig:mlpweigths} shows the models' weights for each architecture under analysis after the learning process. The MLP models incorporate all independent variables with a relevant correlation to the dependent variable, which indicates MLP is acceptable for generating a global model for energy consumption.

\section{Evaluating the proposed models}\label{sec:performance-accuracy-analysis}

In this section, the power consumption models proposed in previous sections are evaluated both in terms of their accuracy and computational cost. For this purpose, different metrics and the 10-fold cross-validation (CV) method were employed. 

\subsection{The accuracy of the models}\label{sec:accuracy}

In order to evaluate the accuracy of each model, we applied four different classes of metrics: scale-dependent, percentage error, relative error, and scale-free error metrics \cite{mape}.

{\it Scale-dependent} metrics are simple to understand and calculate, but cannot be applied to compared models of series with different scales. {\it Percentage error} metrics are scale independent, overcoming the limitations of scale dependent metrics. However, such metrics return infinite or undefined values when zeroes exist within the series. {\it Relative error} metrics are also scale independent metrics but they are restricted to some statistic methods when errors are small. Finally, {\it scale-free error} metrics never provide infinite or undefined values, and they can be applied to compare different estimate methods either over a single or multiple series. 

In this paper, we  used six metrics to evaluate our methods in order to ease future comparison with other methods. The first metric used is Squared Error (SE) given by:

\begin{equation}    \label{eq:se}
        SE_i = (y_i - \hat{y_i})^2
\end{equation}
where $y_i$ is the $i^\textit{th}$ observed value, and $\hat{y}_i$ is the  $i^\textit{th}$ estimate value.

The second metric is the Absolute Error (AE), which is defined by:

\begin{equation}
    AE_i = |y_i - \hat{y_i}|
\label{eq:ae}
\end{equation}

Although these two metrics are scale-dependent, they are used in related literature, such as in \cite{povoa2013model,piga-2013}. The Percentage Error (PE) \cite{mape} is a metric given by the ratio between the difference:

\begin{equation}
PE_i = \nicefrac{(y_i - \hat{y}_i)}{y_i}  
\label{eq:pe}
\end{equation}

In this metric, positive and negative values can cancel each other, leading the average to approach to zero. The Absolute Percentage Error (APE) \cite{mape} is another percentage error metric like PE, except by absolute function as defined by:

\begin{equation}
 APE_i = |\nicefrac{(y_i - \hat{y})} {y_i}|
\label{eq:ape}
\end{equation}

The Absolute Scaled Error (ASE) \cite{mase} is a scale-free error metric, which is frequently used to measure the estimation accuracy. ASE avoids the common problems in conventional accuracy metrics described previously. It is defined as:

%\begin{equation}
%    ASE_i = \frac{(|y_i - \hat y_i|)}{\frac{1}{(n-1)}\sum_{j=2}^{n} |y_j - y_{j-1}|}
%\label{eq:ase}
%\end{equation}
\begin{equation}
    ASE_i = \nicefrac{(|y_i - \hat y_i|)}{\Big(\big(\nicefrac{1}{(n-1)}\big)\sum_{j=2}^{n} |y_j - y_{j-1}|\Big)}
\label{eq:ase}
\end{equation}

where $n$ is the sample size.

For the above mentioned metrics, the closer are the results to zero, the more accurate are the models. Another metric is R$^2$, given by:

\begin{equation}
R^2 = 1 - \nicefrac{\Big(\sum_{i = 1}^{n}(y_i - \hat{y}_i)^2\Big)}{\Big(\sum_{i = 1}^{n}(y_i - \bar{y})^2\Big)}
\label{eq:r2}
\end{equation}

This metric show how well the estimated values produced by a model fit the actual ones. The result lies between 0 and 1, where 0 means the model does not provide any explanation about the data, and 1 refers to a perfect adjust.

\begin{table*}[b!]
    \centering
    \scalefont{1.0}
    \caption{The average and the standard deviation of the Squared Error (SE), Absolute Error (AE), Percentage Error (PE), Absolute Percentage Error (APE), Absolute Scaled Error (ASE), and R$^2$ metrics obtained by the 10-fold Cross-Validation.}
    \resizebox{\textwidth}{!}{
    \begin{tabular}{cc|rrrrr|rrrrr|r}
\hline
    & & \multicolumn{5}{c|}{\textbf{Average}} & \multicolumn{5}{c|}{\textbf{Standard Deviation}} & \\
    \textbf{Arch.} & \textbf{Method}
    & \multicolumn{1}{c}{\textbf{SE}} & \multicolumn{1}{c}{\textbf{AE}} & \multicolumn{1}{c}{\textbf{PE}} & \multicolumn{1}{c}{\textbf{APE}} & \multicolumn{1}{c|}{\textbf{ASE}}
    & \multicolumn{1}{c}{\textbf{SE}} & \multicolumn{1}{c}{\textbf{AE}} & \multicolumn{1}{c}{\textbf{PE}} & \multicolumn{1}{c}{\textbf{APE}} & \multicolumn{1}{c|}{\textbf{ASE}} 
    & \multicolumn{1}{c}{\textbf{R$^2$}} \\
\hline
\multirow{3}{*}{\rotatebox[origin=c]{90}{\textit{A1}}} 
    & \textit{MLR} &  7.2878  & 1.6858 & -0.4008\% & 3.8598\%  & 1.2426 & 56.1252  & 2.1064 & 5.6912\%  & 4.1994\%  & 1.5532 & 96.8650\% \\
    & \textit{RET} & 14.8066  & 2.8970 & -0.7901\% & 6.8144\%  & 2.1249 & 54.7590  & 2.5281 & 8.9543\%  & 5.8610\%  & 1.8534 & 93.6364\% \\
    & \textit{MLP} &  6.1053  & 1.4895 & 0.0332\%  & 3.3382\%  & 1.1040 & 56.3053  & 1.9594 & 5.1738\%  & 3.9961\%  & 1.4517 & 97.3777\% \\
\hline
\multirow{3}{*}{\rotatebox[origin=c]{90}{\textit{A2}}}   
    & \textit{MLR} & 4.9962   & 1.3446 & -0.0078\% & 0.5332\%  & 2.2536 & 21.4396  & 1.7845 & 0.8705\%  & 0.6880\%  & 2.9887 & 91.6575\% \\
    & \textit{RET}  & 4.8958   & 1.3094 & -0.0067\% & 0.5164\%  & 2.2015 & 20.3679  & 1.7828 & 0.8573\%  & 0.6844\%  & 2.9973 & 91.8226\% \\
    & \textit{MLP} & 3.7707   & 1.1169 & -0.0115\% & 0.4424\%  & 1.8725 & 18.0375  & 1.5873 & 0.7533\%  & 0.6115\%  & 2.6572 & 93.7082\% \\
\hline
\multirow{3}{*}{\rotatebox[origin=c]{90}{\textit{Mix}}}  
    & \textit{MLR} &  14.0595 & 2.6209 & -0.4336\% &  3.7864\% & 2.5888 &  53.7892 & 2.6807 &  5.9828\% &  4.6517\% & 2.6474 & 99.8654\% \\
    & \textit{RET}  & 150.7207 & 7.7339 & -4.1102\% & 11.7914\% & 7.6287 & 533.8658 & 9.5315 & 19.3902\% & 15.9341\% & 9.4011 & 98.5565\% \\
    & \textit{MLP} &   6.3264 & 1.5471 & -0.3946\% &  2.3506\% & 1.5232 &  48.7559 & 1.9796 &  4.3027\% &  3.6313\% & 1.9479 & 99.9394\% \\
\hline
\end{tabular}
    }
    \label{tab:accuracy}
\end{table*}

The metrics described above provide values closer to $0.00$ for better models and far from $0.00$ for worse ones. The PE metric can provide either negative and positive values, and the remainder metrics result only in positive values.

Next, the accuracy of each proposed model is evaluated applying the 10-fold cross validation method \cite{cv}. For each test, the estimated value for the power consumption is compared to the actual measured value. Table \ref{tab:accuracy} presents the average and standard deviation for the six metrics. All models presented $R^2 > 91\%$. In particular, MLR models have low average errors for all metrics considering A1 and A2 architectures. However, when MLR is applied to fit the mix of architectures into a unique model, the error increases significantly. A similar effect occurs with RET models, whose accuracy is even worse than MLR models for the mix of architectures.

MLP models presented the best accuracy from the experiments. However, MLR are simpler and less costly models whose accuracy approach the MLP's accuracy. It suggests non-linear relations with low significance between the independent variables and the dependent one. This evidence is supported by the the average and standard deviation, which are close but not equal.

\begin{figure*}[t!]
\includegraphics[width=\textwidth]{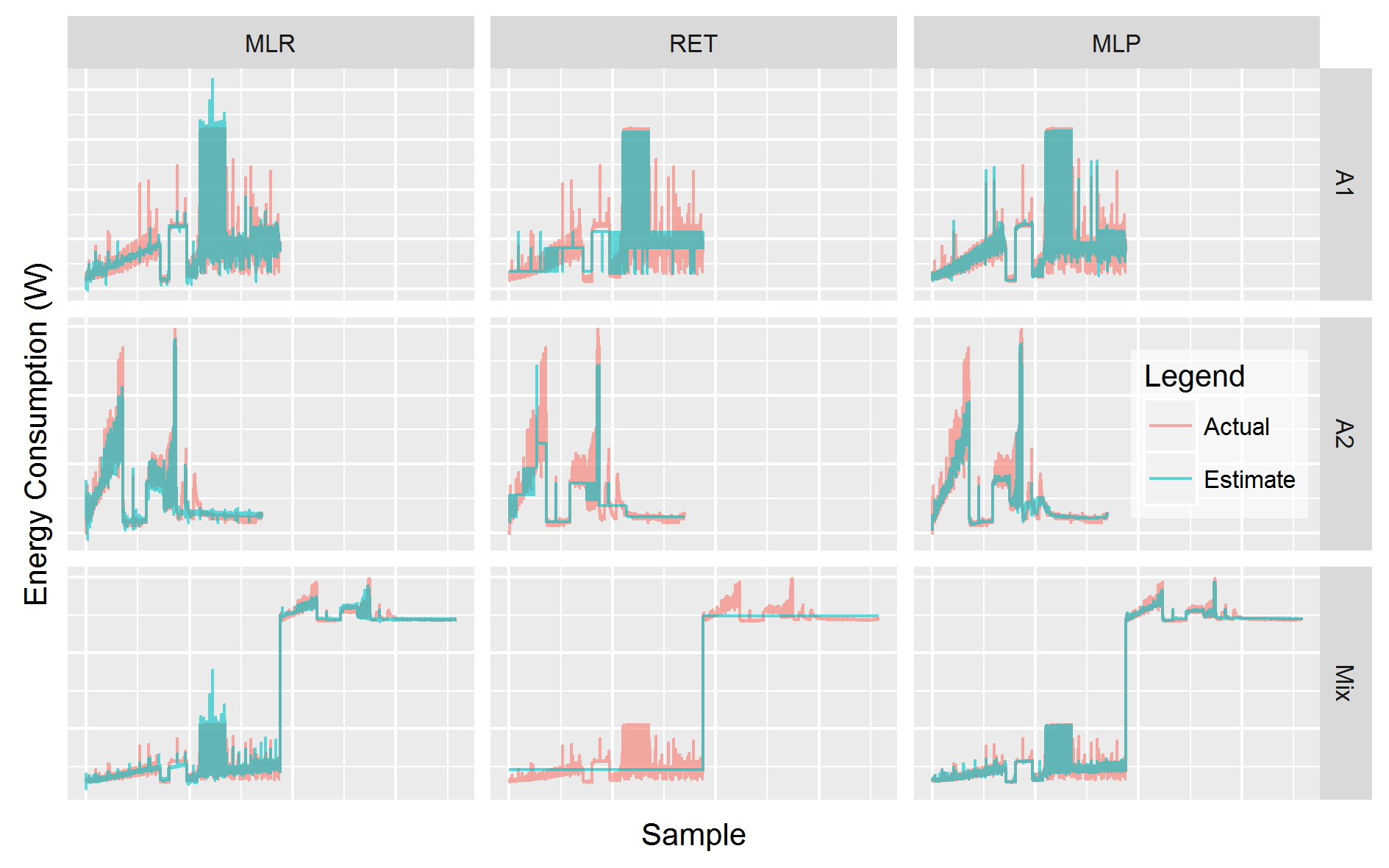}
\caption{Comparison between actual and estimated values.}
\label{fig:models_accuracy}
\end{figure*}

Noticeably, RET models present the worst accuracy from the three models. This can be explained as RET clusters data before estimating the power consumption. Indeed, power consumption cannot be explained for a small subset of dependent variables. However, all of the variables provide enough information for estimating power consumption, which 
hinders the clustering.

As depicted in Figure~\ref{fig:models_accuracy}, the estimated and actual power consumption are close. Considering the results, we can conclude the MLP models provide better estimations for power consumption, while MLR are simpler models which present similar performance in terms of accuracy. Furthermore, experimental results also show that RET models do not provide accurate estimations for power consumption when compared to MLP and MLR models, mainly when dealing with mixed architectures in the same estimator.

\subsection{Computational cost}\label{sec:performance}

Besides accuracy, computational cost to estimate power consumption is also assessed. All the three methods have a training phase and the application phase. During the training phase the method is fed with traces of operating system variables and actual power consumption reads from a meter, and calibrates its internal parameters to produce accurate estimates. 
Usually, the training phase is far more computationally expensive than the application phase. Table~\ref{tab:performance-times} shows the execution times (in minutes) for the training phase of each method. Executoin times were obtained with the GNU \textit{time} tool~\cite{time}. 

MLR presents the fastest training phase, around $48.72$ minutes. Not surprisingly, MLP is the slowest one (around 6.7 days). In general, such models will be trained once for each particular architecture profile. Machines with similar architecture can reuse the same model.

\begin{table}[h!]
\caption{Execution times for the training phase of each method (in minutes). }
\centering
\scalefont{1.0}
\begin{tabular}{c|rrrrrr}
\cline{1-4}
& \multicolumn{1}{c}{\textbf{A1}} & \multicolumn{1}{c}{\textbf{A2}} & \multicolumn{1}{c}{\textbf{Mix}} \\
\cline{1-4}
\textbf{MLR} &    0.81 &    0.65 &    2.69    \\
\textbf{RET} &    5.48 &    5.50 &    5.58    \\
\textbf{MLP} & 3,693.26 & 2,446.07 & 9,738.97 \\
\hline
\end{tabular}

\label{tab:performance-times}
\end{table}

During the application phase, the model fed with an array $V$ of measures of resources utilization collected from the operating system and produces an estimate of power consumption. This is a computationally low cost operation that requires a few dozens of floating-point operations. Depending on the frequency that such calculation need to be executed (e.g. once a second), its computational cost may be negligible (far less than 1\% of cpu utilization).

\section{Related Work}\label{sec:related_work}

A large number of papers has been published on modeling computers power consumption, including some surveys  \cite{reda2012power,orgerie2014survey,model_survey}. 
Several models have been proposed to estimate the energy consumption of processors \cite{joseph2001run,contreras2005power,bertran2013systematic,bertran2010decomposable}. Most of them consist of linear regression-based models which are fed with hardware performance counters. In \cite{isci2003runtime}, a model was proposed combining real total power measurement with hardware counters measurement to estimate per-component energy consumption. Our approach is different from those works because our goal is to estimate energy consumption for the entire machine, not limited to the processor.

Other papers address the modeling of the entire computer (e.g., from commodity computers to datacenter servers) proposing linear models composed by the summation of the energy consumed by its subcomponents \cite{lewis2008run,orgerie2012energy,xiao2013virtual,lent2013model}. For instance, Lewis et al. \cite{lewis2008run} propose an aggregated model which considers CPU, memory, electromechanical components, peripherals, and hard disk consumption. The models have coefficients for each component that are adjusted using linear regression. 
The energy consumed by virtual machines is also modeled in \cite {xiao2013virtual}. Other non linear models are also proposed for modeling the entire computer energy consumption \cite{fan2007power,tang2011dynamic}. Our work is different as our objective is not to model energy consumption of the computer as an explicit summation of the consumption of its subcomponents. Instead, our models are fed with system variables carefully selected (by their ability to explain the model) in order to estimate energy consumption with high accuracy. Also, our work propose and compare models based on three different techniques.

As mentioned, regression models are numerous for modeling energy consumption. For instance, Piga~\cite{piga-2013} defined a global center-level approach to power and performance optimization for Web Server Datacenters. Their model is based on linear and non-linear regression techniques, while using the k-means to identify non-linear correlation and the Correlation-based Feature Selection (CFS) for removing independents variables that do not provide significant explanation for the power consumption. Our focus, instead is to model individual computers based on observable operating system measures.

Da Costa et al.~\cite{da2010methodology} modeled computer energy consumption based on performance counters provided by two tools ({Linux \tt pidstat} and {\tt collectd}). The paper describes the methodology for reducing from a set of 165 explanatory variables to a small number of variables which can explain the model with high accuracy. The model is intended to estimate energy consumption at process level. Our work is different regarding the variety of techniques used, and modeling the whole machine energy consumption.

Comparing our models with the best related work results, considering the absolute percentage error, our best MLP specific and global models presented an error rate of 2.35\%, and 0.44\%, correspondingly, while \cite{lewis2008run,piga-2013,da2010methodology,dhiman} have an error rate of $\sim 4.0\%$, $\sim 4.4\%$, $\sim 10.0\%$, and $\sim 6.0\%$, respectively.

We did not compare our models against all related work once some models have no direct correlation with our approach, measuring, for example, energy consumption of a specific hardware component or a whole data center, or providing an abstract model evaluation.

%Table~\ref{tab:models-comparison} indicates that our models outperforms current techniques to power consumption estimation.

%\begin{table}[hbt]
%    \centering
%    \caption{Evaluation of Different Power Consumption Models}
%    \label{tab:models-comparison}
%     \resizebox{\columnwidth}{!}{
%    \input{tables/performance.comparison}}
%\end{table}

By the best of our knowledge, our work provides the following novel contribution: ($i$) propose and compare three different models to estimate the power consumption for more than one hardware configuration; ($ii$) employ the MIC method to analysis the correlation between independent variables and the power consumption variable; ($iii$) the proposed models are fed with commodity system variables commonly provided by Linux, for better portability; ($iv$) accuracy analysis using several metrics along with cross-validation in order to verify precision and overfitting issues.

\section{Conclusion 
%and Future Work
}\label{sec:conclusion}

This paper presented a characterization of energy consumption of entire machines based on resources utilization variables. Experiments were carried out using synthetic workloads in order to discover what resources and modeling methods present higher correlation to energy consumption. We show that it is possible to estimate energy consumption  by sampling  variables provided by common operating systems and employing MLR, RET, or MLP methods. We proposed nine models that provide accurate estimation on energy consumption with an accuracy of 99.9\%, and average squared error of $6.32$ watts with standard deviation of $48.76$. 

All models evaluated can be fully implemented in software, providing a cost-effective mechanism for estimating energy consumption. Such models can be deployed from a single machine to all machines in a large datacenter at no cost and negligible overhead. Our proposal can be useful for several aims, e.g., to provide instant information on energy consumption in a per machine basis in a datacenter. Such information can be consumed by management tools to provide accurate views of energy efficiency of the datacenter. Also, resource schedulers and consolidation tools could benefit from our proposal on different environments such as high performance clusters and cloud infrastructures.

%As future work, we intend to build a database of energy consumption models for different hardware characteristics, and develop energy-based scheduling algorithms for the bag-of-tasks paradigm in order to provide energy-efficient allocation into a distributed environment. We also intend to apply Deep Learning to verify the possible accuracy improvement.

\begin{acknowledgements}
Authors thank CAPES and RNP for partially supporting this research project. Hermes Senger thanks CNPq (Contract Number 305032/2015-1) for their support. The authors declare that no funding body played any role in the design or conclusion of this study.
\end{acknowledgements}

\bibliographystyle{spphys}
\bibliography{references} 

\begin{thebibliography}{10}
\providecommand{\url}[1]{{#1}}
\providecommand{\urlprefix}{URL }
\expandafter\ifx\csname urlstyle\endcsname\relax
  \providecommand{\doi}[1]{DOI \discretionary{}{}{}#1}\else
  \providecommand{\doi}{DOI \discretionary{}{}{}\begingroup
  \urlstyle{rm}\Url}\fi

\bibitem{gartner_co2}
Gartner, Gartner energy \& utilities it summit 2007: an invitation only event.
\newblock Tech. rep., Gartner, Dallas, EUA (2007)

\bibitem{greenpeace}
G.~Cook, How clear is your cloud? catalysing an energy revolution.
\newblock Tech. rep., Greenpeace International, Amsterdam, The Netherlands
  (2012).
\newblock \urlprefix\url{http://goo.gl/yd1FAS}

\bibitem{gao2012s}
P.X. Gao, A.R. Curtis, B.~Wong, S.~Keshav, in \emph{Proceedings of the ACM
  SIGCOMM 2012 Conference on Applications, Technologies, Architectures, and
  Protocols for Computer Communication} (ACM, New York, NY, USA, 2012), SIGCOMM
  '12, pp. 211--222

\bibitem{koomey2011growth}
J.~Koomey, \emph{Growth in data center electricity use 2005 to 2010}, vol.~1
  (Analytics Press, Oakland, CA, 2011)

\bibitem{orgerie_et_al_2010}
A.C. Orgerie, M.~Assunção, L.~Lefèvre, in \emph{Grids, Clouds and
  Virtualization}, ed. by M.~Cafaro, G.~Aloisio, Computer Communications and
  Networks (Springer London, 2011), pp. 143--166

\bibitem{lee_e_zomaya_2012}
Y.~Lee, A.~Zomaya, The Journal of Supercomputing \textbf{60}, 268 (2012)

\bibitem{song_et_al_2009}
Y.~Song, H.~Wang, Y.~Li, B.~Feng, Y.~Sun, in \emph{CCGRID \'09.} (2009), pp.
  148 --155

\bibitem{barroso_e_holzle_2007}
L.A. Barroso, U.~H{\"o}lzle, Computer \textbf{40}(12), 33 (2007)

\bibitem{song2013simplified}
S.~Song, C.~Su, B.~Rountree, K.W. Cameron, in \emph{Parallel \& Distributed
  Processing (IPDPS), 2013 IEEE 27th International Symposium on} (IEEE, 2013),
  pp. 673--686

\bibitem{naveh2011power}
A.~Naveh, D.~Rajwan, A.~Ananthakrishnan, E.~Weissmann, in \emph{Hot Chips}
  (2011)

\bibitem{contreras2005power}
G.~Contreras, M.~Martonosi, in \emph{Low Power Electronics and Design, 2005.
  ISLPED'05. Proceedings of the 2005 International Symposium on} (IEEE, 2005),
  pp. 221--226

\bibitem{joseph2001run}
R.~Joseph, M.~Martonosi, in \emph{Proceedings of the 2001 international
  symposium on Low power electronics and design} (ACM, 2001), pp. 135--140

\bibitem{isci2003runtime}
C.~Isci, M.~Martonosi, in \emph{Proceedings of the 36th annual IEEE/ACM
  International Symposium on Microarchitecture} (IEEE Computer Society, 2003),
  p.~93

\bibitem{bertran2010decomposable}
R.~Bertran, M.~Gonzalez, X.~Martorell, N.~Navarro, E.~Ayguade, in
  \emph{Proceedings of the 24th ACM International Conference on Supercomputing}
  (ACM, 2010), pp. 147--158

\bibitem{povoa2013model}
L.~Venezian~Povoa, P.~Bignatto~Junior, C.~Monteiro, D.~Mueller, C.~Marcondes,
  H.~Senger, in \emph{IEEE Symposium on Computers and Communications (ISCC)}
  (IEEE Computer Society, 2013), pp. 1--6

\bibitem{iaee}
P.M. Dusso, A monitoring system for wattdb: An energy-proportional database
  cluster.
\newblock Graduation thesis, Informatics Institute, Federal University of Rio
  Grande do Sul, Porto Alegre, Brazil (2012).
\newblock \urlprefix\url{http://goo.gl/8gheFW}

\bibitem{iaee_impl}
P.M. Dusso.
\newblock energyagent (2012).
\newblock \urlprefix\url{https://github.com/pmdusso/energyagent}

\bibitem{mine}
D.N. Reshef, Y.A. Reshef, H.K. Finucane, S.R. Grossman, G.~McVean, P.J.
  Turnbaugh, E.S. Lander, M.~Mitzenmacher, P.C. Sabeti, Science
  \textbf{334}(6062), 1518 (2011)

\bibitem{pazzani1999independent}
M.J. Pazzani, S.D. Bay, in \emph{Proceedings of the Twenty-First Annual Meeting
  of the Cognitive Science Society} (1999), pp. 525--530

\bibitem{da2010methodology}
G.~Da~Costa, H.~Hlavacs, in \emph{2010 11th IEEE/ACM International Conference
  on Grid Computing} (IEEE, 2010), pp. 290--297

\bibitem{stress}
A.~Waterland.
\newblock stress (2014).
\newblock \urlprefix\url{http://people.seas.harvard.edu/~apw/stress/}

\bibitem{cpulimit}
A.~Marletta.
\newblock cpulimit (2012).
\newblock \urlprefix\url{https://github.com/opsengine/cpulimit}

\bibitem{iperf}
{NLANR/DAST}.
\newblock Iperf - the tcp/udp bandwidth measurement tool (2011).
\newblock \urlprefix\url{https://iperf.fr/}

\bibitem{2375a10}
{Yokogawa Electric Corporation}, \emph{0.5 Class Transducer for Power
  Application} (2009).
\newblock \urlprefix\url{http://goo.gl/f43ZwV}

\bibitem{mw100}
{Yokogawa Electric Corporation}, \emph{Data Acquisition Unit MW100} (2013).
\newblock \urlprefix\url{http://goo.gl/7mcNV8}

\bibitem{ks-test}
G.~Marsaglia, W.W. Tsang, J.~Wang, Journal of Statistical Software
  \textbf{8}(18), 1 (2003)

\bibitem{clt}
J.R. Rice, \emph{Mathematical Statistics and Data Analysis} (Thomson
  Books/Cole, United States of America, 2007), 3rd edn., chap. 5 - Limit
  Theorems

\bibitem{CART}
L.~Breiman, J.~Friedman, C.J. Stone, R.~Olshen, \emph{Classification and
  Regression Trees} (Chapman and Hall, New York, USA, 1984)

\bibitem{intro_rpart}
T.M. Therneau, E.J. Atkinson, An introduction to recursive partitioning using
  the rpart routines.
\newblock Tech. rep., The R Foundation (2015).
\newblock \urlprefix\url{https://goo.gl/YmMfNb}

\bibitem{mlp}
S.~Haykin, \emph{Neural Networks: A Comprehensive Foundation}, 2nd edn.
  (Prentice Hall, 1998)

\bibitem{mlp_biometric}
V.B. Semwal, M.~Raj, G.~Nandi, Robotics and Autonomous Systems \textbf{65}, 65
  (2015).
\newblock \doi{10.1016/j.robot.2014.11.010}

\bibitem{mlp_thermal}
M.~De~Lozzo, P.~Klotz, B.~Laurent, Engineering Applications of Artificial
  Intelligence \textbf{26}(10), 2270 (2013).
\newblock \doi{10.1016/j.engappai.2013.07.001}

\bibitem{mlp_ocean}
A.~Altunkaynak, Ocean Engineering \textbf{58}, 144 (2013).
\newblock \doi{10.1016/j.oceaneng.2012.08.005}

\bibitem{mlp_climatology}
R.~Velo, P.~L{\'o}pez, F.~Maseda, Energy Conversion and Management \textbf{81},
  1 (2014).
\newblock \doi{10.1016/j.enconman.2014.02.017}

\bibitem{minibatch}
M.~Li, T.~Zhang, Y.~Chen, A.J. Smola, in \emph{Proceedings of the 20th ACM
  SIGKDD International Conference on Knowledge Discovery and Data Mining} (ACM,
  New York, NY, USA, 2014), KDD '14, pp. 661--670.
\newblock \doi{10.1145/2623330.2623612}

\bibitem{mape}
R.J. Hyndman, A.B. Koehler, International Journal of Forecasting pp. 679--688
  (2006)

\bibitem{piga-2013}
L.~de~Paula Rosa~Piga, Modeling, characterization, and optimization of web
  server power in data centers.
\newblock Ph.D. thesis, Institute of Computing (IC), University of Campinas
  (UNICAMP), Campinas, Brazil (2013)

\bibitem{mase}
R.J. Hyndman, Foresight: The International Journal of Applied Forecasting (4),
  43 (2006)

\bibitem{cv}
R.~Kohavi, in \emph{Proceedings of the 14th International Joint Conference on
  Artificial Intelligence - Volume 2} (Morgan Kaufmann Publishers Inc., San
  Francisco, CA, USA, 1995), IJCAI'95, pp. 1137--1143

\bibitem{time}
D.~MacKenzie, \emph{GNU time 1.7} (2010).
\newblock \urlprefix\url{http://goo.gl/QYPqN3}

\bibitem{reda2012power}
S.~Reda, A.N. Nowroz, Foundations and Trends in Electronic Design Automation
  \textbf{6}(2), 121 (2012)

\bibitem{orgerie2014survey}
A.C. Orgerie, M.D.d. Assuncao, L.~Lefevre, ACM Computing Surveys (CSUR)
  \textbf{46}(4), 47 (2014)

\bibitem{model_survey}
M.~Dayarathna, Y.~Wen, R.~Fan, IEEE Communications Surveys Tutorials
  \textbf{18}(1), 732 (2016).
\newblock \doi{10.1109/COMST.2015.2481183}

\bibitem{bertran2013systematic}
R.~Bertran, M.~Gonzalez, X.~Martorell, N.~Navarro, E.~Ayguade, IEEE
  Transactions on Computers \textbf{62}(7), 1289 (2013)

\bibitem{lewis2008run}
A.W. Lewis, S.~Ghosh, N.F. Tzeng, HotPower \textbf{8}, 17  (2008)

\bibitem{orgerie2012energy}
A.C. Orgerie, L.~Lef{\`e}vre, I.~Gu{\'e}rin-Lassous, The Journal of
  Supercomputing \textbf{62}(3), 1139 (2012)

\bibitem{xiao2013virtual}
P.~Xiao, Z.~Hu, D.~Liu, G.~Yan, X.~Qu, Journal of Network and Computer
  Applications \textbf{36}(2), 818 (2013)

\bibitem{lent2013model}
R.~Lent, Sustainable Computing: Informatics and Systems \textbf{3}(2), 80
  (2013)

\bibitem{fan2007power}
X.~Fan, W.D. Weber, L.A. Barroso, in \emph{ACM SIGARCH Computer Architecture
  News} (ACM, 2007), pp. 13--23

\bibitem{tang2011dynamic}
C.J. Tang, M.R. Dai, in \emph{System Integration (SII), 2011 IEEE/SICE
  International Symposium on} (IEEE, 2011), pp. 1159--1164

\bibitem{dhiman}
G.~Dhiman, K.~Mihic, T.~Rosing, in \emph{Design Automation Conference (DAC),
  2010 47th ACM/IEEE} (2010), pp. 807 -- 812.
\newblock \doi{10.1145/1837274.1837478}

\end{thebibliography}

\end{document}